\pdfoutput=1

\documentclass[11pt]{article}

\usepackage[]{ACL2023}

\usepackage{times}
\usepackage{latexsym}
\usepackage{graphicx} 

\usepackage{pgfplots}
\pgfplotsset{width=10cm,compat=1.9}
\usepackage{tikz}
\usepackage{subcaption}
\usepackage{fontawesome5}
\usepackage{tikz-cd}
\usepackage[dvipsnames]{xcolor}
\usepackage{booktabs}

\usepackage{soul}
\usepackage[T1]{fontenc}

\usepackage[utf8]{inputenc}

\usepackage{microtype}

\usepackage{inconsolata}
\usepackage{xcolor}

\usepackage{hyperref}

\tikzset{
    lm-bg halfcircle/.style={%
        mark=halfcircle*,
        mark color=teal!75!white,
        every mark/.append style={rotate=#1}
    }
}
\tikzset{
    mt-bg halfcircle/.style={%
        mark=halfcircle*,
        mark color=Salmon!75!white,
        every mark/.append style={rotate=#1}
    }
}

\newcommand{\bs}[1]{\textcolor{black}{#1}}

\newcommand{\ar}[1]{\textcolor{black}{#1}}

\usepackage{todonotes}
\usepackage{soul} 
\usepackage{marginnote}
\usepackage{url}

\usepackage[capitalize]{cleveref}

\usepackage{listings}
\definecolor{TodoColor}{rgb}{1,0.7,0.6}
\definecolor{TodoColor2}{rgb}{0.7,0.7,0.9}
\definecolor{TodoColor3}{rgb}{0.5,0.8,0.5}

\title{Translation in the Hands of Many: \\
Machine Translation as a (Lay) User-Facing Technology}

\title{Translation in the Hands of Many: \\
Centering Lay Users in Machine Translation Interactions}

\author{Beatrice Savoldi, Alan Ramponi, Matteo Negri, Luisa Bentivogli
         \\ Fondazione Bruno Kessler, Italy \\
         \texttt{\{bsavoldi,alramponi,negri,bentivo\}@fbk.eu}}

\begin{document}
\maketitle
\begin{abstract}

\bs{Converging societal and technical factors have transformed language technologies into user-facing applications used by the general public across languages. Machine Translation (MT) has become a global tool, with cross-lingual services now also supported by dialogue systems powered by multilingual Large Language Models (LLMs). Widespread accessibility has extended MT’s reach to a vast base of \textit{lay~users}, many with little to no expertise in the languages or the technology itself. And yet, 
the understanding of MT consumed by such a diverse group of 
users---their needs, experiences, and interactions with multilingual systems---remains limited.
In our position paper, we first trace the evolution of MT user profiles, focusing on non-experts and how their engagement with technology may shift with the rise of LLMs. Building on an interdisciplinary body of work, we identify three factors---usability, trust, and literacy---that are central to shaping user interactions and must be addressed to align MT with user needs. By examining these dimensions, we provide insights to guide the progress of more user-centered MT.}

\end{abstract}

\section{Introduction}


The success of technology hinges on its ability to serve users, and Natural Language Processing (NLP) confronts this challenge as it transitions from an  
academic pursuit to a set of 
impactful tools. Among them, MT stands out as a cornerstone 
application,
with current breadth and quality that  
fostered wider adoption \citep{WANG2022143}. 
Multilingual demands \citep{moorkens2024artificial},
paired with the accessibility of online systems, has put MT at the forefront of user-facing language technologies.
%
%
%
Once confined to professional 
settings, MT is now used by millions 
\citep{pitman2021translate}, bringing into its fold an 
array of \textit{lay users} in contexts ranging from casual interactions 
\citep{gao}
to critical domains such as healthcare and employment \citep{patil2014use, dew2018development, liebling-etal-2022-opportunities, valdez2023migrant}.

\begin{figure}[!t]
\begin{subfigure}{.24\textwidth}
    \resizebox{\linewidth}{!}{%

    \begin{minipage}[t]{.99\linewidth}
    \centering
    \strut\vspace*{-9mm}\newline
    
        \begin{tikzpicture}
            \begin{axis}[
                /pgf/number format/1000 sep={},
                ylabel={Percentage (\%)},
                ylabel style={yshift=-1mm},
                xmin=2015, xmax=2024,
                ymin=0, ymax=60,
                xtick={2015,2016,2017,2018,2019,2020,2021,2022,2023,2024,2025},
                ytick={60,50,40,30,20,10,0},
                height=3.6cm,
                width=4.6cm,
                legend pos=north west,
                ymajorgrids=true,
                grid style=dashed,
                title style={yshift=-3mm},
                xlabel style = {font=\footnotesize},
                ylabel style = {font=\footnotesize},
                xticklabel style = {font=\scriptsize, rotate=90},
                yticklabel style = {font=\scriptsize},
                legend style={font=\scriptsize},
            ]

            \addplot[
                color=Salmon!75!white,
                mark=*,
                draw=black!75!white,
                ]
                coordinates {
                (2015,11.96)(2016,12.71)(2017,13.51)(2018,14.21)(2019,15.12)(2020,15.09)(2021,14.29)(2022,12.90)(2023,12.20)(2024,10.48)
                };

            \addplot[
                color=teal!75!white,
                mark=*,
                draw=black!75!white,
                ]
                coordinates {
                (2015,1.19)(2016,2.36)(2017,2.91)(2018,3.86)(2019,7.31)(2020,11.48)(2021,16.25)(2022,22.28)(2023,34.81)(2024,56.94)
                };

            \addplot[
                color=Salmon!75!white,
                lm-bg halfcircle=90,
                draw=black!75!white,
                ]
                coordinates {
                (2015,0.17)(2016,0.49)(2017,0.33)(2018,0.69)(2019,0.99)(2020,1.31)(2021,1.43)(2022,1.76)(2023,2.61)(2024,4.41)
                };
            \end{axis}
        \end{tikzpicture}

        \end{minipage}
}%

\end{subfigure}%
\hfill
\begin{subfigure}{.24\textwidth}

    \resizebox{\linewidth}{!}{%

    \begin{minipage}[t]{.99\linewidth}
    \centering
    \strut\vspace*{-9mm}\newline
    
        \begin{tikzpicture}
            \begin{axis}[
                /pgf/number format/1000 sep={},
                ylabel={\textcolor{white}{\%}},
                ylabel style={yshift=-3mm},
                xmin=2015, xmax=2024,
                ymin=0, ymax=10,
                xtick={2015,2016,2017,2018,2019,2020,2021,2022,2023,2024,2025},
                ytick={10,5,0},
                height=3.6cm,
                width=4.6cm,
                legend pos=north west,
                ymajorgrids=true,
                grid style=dashed,
                title style={yshift=-3mm},
                xlabel style = {font=\footnotesize},
                ylabel style = {font=\footnotesize},
                xticklabel style = {font=\scriptsize, rotate=90},
                yticklabel style = {font=\scriptsize},
                legend style={font=\scriptsize},
            ]

            \addplot[
                color=Goldenrod!75!white,
                mark=*,
                draw=black!75!white,
                ]
                coordinates {
                (2015,1.40)(2016,5.59)(2017,7.24)(2018,7.07)(2019,8.22)(2020,9.14)(2021,8.24)(2022,8.79)(2023,8.94)(2024,9.99)
                };

            \addplot[
                color=Goldenrod!75!white,
                mt-bg halfcircle=270,
                draw=black!75!white,
                ]
                coordinates {
                (2015,0.17)(2016,0.75)(2017,0.64)(2018,0.63)(2019,0.62)(2020,0.74)(2021,0.92)(2022,0.75)(2023,0.89)(2024,0.65)
                };

            \addplot[
                color=Goldenrod!75!white,
                lm-bg halfcircle=270,
                draw=black!75!white,
                ]
                coordinates {
                (2015,0)(2016,0.12)(2017,0.06)(2018,0.15)(2019,0.41)(2020,0.69)(2021,0.96)(2022,1.29)(2023,2.94)(2024,6.13)
                };
            \end{axis}
        \end{tikzpicture}

        \end{minipage}
}%
\end{subfigure}%

\caption{Trend of interest in \emph{machine translation} \colorbox{Salmon!75!white}{\textsc{\textbf{mt}}}, \emph{language models} \colorbox{teal!75!white}{\textsc{\textbf{\textcolor{white}{lm}}}}, \emph{users} \colorbox{Goldenrod!75!white}{\textsc{\textbf{u}}}, and combinations thereof in the ACL community over the last 10 years. Besides illustrating the rapid growth of LLM studies, \bs{the left panel highlights the increase in MT research incorporating LLMs (\colorbox{Salmon!75!white}{\textsc{\textbf{mt}}}+\colorbox{teal!75!white}{\textsc{\textbf{\textcolor{white}{lm}}}}), while the right panel shows rising attention to users, particularly in LLM-related work (\colorbox{teal!75!white}{\textsc{\textbf{\textcolor{white}{lm}}}}+\colorbox{Goldenrod!75!white}{\textsc{\textbf{u}}}).}\footnotemark}

\label{fig:trend-overview}

\end{figure}

Despite MT’s broad reach and potential for social impact in sensitive scenarios \citep{vieira2021understanding}, still little is known about its evolving relationship with the general public, how non-expert  
users interact with it, or how it caters to their needs \citep{carpuat2025interdisciplinaryapproachhumancenteredmachine}. MT research has mainly focused on modeling advancements and---although translation studies have called for greater attention to end-user perspectives \citep{guerberof2023ethics} \bs{and related efforts from human-computer interaction \citep{zhang, global-meeting}}---MT works that actively involve lay people and 
their experiences are  still 
rare \citep{mehandru-etal-2023-physician, briakou-etal-2023-explaining}.
\footnotetext{\bs{Details on the ACL Anthology queries are provided in Appendix~\ref{app:acl-query}. For a complementary view, Figure~\ref{fig:trend-overview-2} in the appendix also shows absolute counts in the trends of interest over the last ten years.}}

In the wake of broader calls to bridge MT \citep{liebling2021three} and language technologies with user-centered research \citep{heuer2021methods, kotnis2022human}, we posit that it is time to fill this gap and focus on how to support interactions between systems and lay people. 
Arguably, the rise of powerful, instruction-following 
LLMs 
\citep[\textit{inter alia}]{touvron2023llama, achiam2023gpt, geminiteam2024geminifamilyhighlycapable, üstün2024ayamodelinstructionfinetuned}
engaging non-experts via chat interfaces has heightened user concerns (see Figure~\ref{fig:trend-overview}, 
\colorbox{teal!75!white}{\textsc{\textbf{\textcolor{white}{lm}}}}+\colorbox{Goldenrod!75!white}{\textsc{\textbf{u}}}
on the \emph{right}) 
and underscores 
the urgency to align with real-world interactions \citep{haque-etal-2022-pixie, liao2023aitransparencyagellms, szymanski2024comparingcriteriadevelopmentdomain}.\footnote{\bs{For wider initiatives towards human-centered approaches in the *CL community, we notice the introduction of the \href{https://2023.emnlp.org/calls/main_conference_papers/}{Human-Centered NLP track} since 2023, as well  the HCI+NLP workshop~\citep{hcinlp-ws-2024-bridging} and the tutorial on Human-Centered Evaluation of Language Technologies~\citep{blodgett-etal-2024-human}.}} 
As MT moves towards LLM-based solutions (see Figure \ref{fig:trend-overview}, 
\colorbox{Salmon!75!white}{\textsc{\textbf{mt}}}+\colorbox{teal!75!white}{\textsc{\textbf{\textcolor{white}{lm}}}}
\emph{vs} 
\colorbox{Salmon!75!white}{\textsc{\textbf{mt}}}
on the \emph{left}), these have the potential to redefine how people engage with multilingual systems, challenging traditional task divisions with new paradigms for cross-lingual communication \citep{ouyang2023shifted, lyu2024paradigm}. 

To set the stage for this shift towards lay users' perspective, we 
examine the evolution of MT from professional settings to its 
wide general adoption (\S\ref{sec:mt-evolution}). We then identify three key factors---\textit{usability~\faHandPointer}, \textit{trust \faCheck}, and \textit{literacy} \faBookOpen---to ground user interactions with automatic translation tools (\S\ref{sec:factors}). Through this lens, we take stock of the current landscape 
to guide MT research in tandem with users (\S\ref{sec:recommendations}).

\bs{We release a curated list of the works discussed in the paper at: \url{https://github.com/hlt-mt/awesome-human-centered-MT}.}



%
%
%


\section{MT and User Evolution} \label{sec:mt-evolution}

Although online systems have existed for some time \citep{yang1998systran, mccarthy2004does, somers2005round}, we are now seeing unprecedented volumes of unrevised MT outputs being directly consumed by the public.\footnote{See the rising volume of Google Translate app downloads and words translated with it \citep{pitman2021translate}.}
\textbf{Historically, real-world applications of MT often regarded so-called ``mixed MT'' workflows }\citep{wagner1983rapid}, where human intervention serves to revise---i.e. post-edit \citep{li2023post}---MT to produce a reliable final translation.
Attention to this usage scenario \citep{church1993good} is reflected in MT
development \citep{green2014human, bentivogli2015evaluation, daems2019interactive}, 
interfaces~\citep{vieira2011review, vela2019improving}, and evaluation 
\citep{popovic2011towards, bentivogli2016neural} using \textit{professional} translators as a target. 
Such a trajectory was also paired with 
empirical experiments on when MT could support \citep{koponen2016machine, moorkens2017assessing} or interfere \citep{federico2014assessing, daems2017identifying} with translators' activity.
%

The advent of stronger models with expanded language coverage---along with the rise of the Web and 
personal devices---progressively altered  
this landscape.
\textbf{MT consumption has now reached wider adoption by the general public, who directly accesses \textit{raw} MT output} in many diverse scenarios,\footnote{Also leading to a decrease in the demand for language skills and professional work \citep{frey2025lost}.} e.g.
to gist content, for multilingual conversations \citep{conversation, pombal2024context}, in education \citep{yang2021measuring, yang2024understanding}, but also in high-stakes domains such as  healthcare \citep{khoong2019assessing, valdez2025google}, migration 
\citep{liebling-etal-2022-opportunities}, and emergency services \citep{TURNER2015136}.\footnote{e.g. with COVID to compensate for interpreters shortages \citep{khoong2022research,anastasopoulos2020tico}.} 
\textbf{This shift} \textbf{to unmediated MT} 
\textbf{has led to a vast, heterogeneous base of \textit{lay users}}
\textbf{and, with it, novel desiderata and concerns}. For one, since lay users may have limited to no proficiency in at least one of the involved languages,\footnote{e.g. the \emph{source} language in gisting and the \emph{target} in communication contexts. See also \citet{nurminen-papula-2018-gist}.} they are more vulnerable to errors. 
Mistranslations can lead to discomfort, misunderstandings, and even life-threatening errors \citep{taira2021pragmatic} and arrests \citep{guardian2017facebook}.
Besides, non-experts 
can have requirements and expectations of which little is known, and 
that cannot be directly informed by existing research on professionals, as shown in the context of LLMs---e.g. \citet{szymanski2024comparingcriteriadevelopmentdomain}, see also Figure \ref{tab:user_comparison}, Appendix \ref{app:user}). 

Indeed, 
general-purpose LLMs are calling for more considerations of users and real-world contexts of use, as demonstrated by surveys to understand how people interact with technologies, for which purposes and needs~\citep[\textit{inter alia}]{tao-etal-2024-chatgpt, Skjuve_Brandtzaeg_Følstad_2024, Kim_2024, STOJANOV2024100243, bodonhelyi2024userintentrecognitionsatisfaction, wang2024user, HYUNBAEK2023102030}. \textbf{Chat-based LLMs} have drawn in millions of users,\footnote{According to \href{https://www.reuters.com/technology/artificial-intelligence/openai-says-chatgpts-weekly-users-have-grown-200-million-2024-08-29/}{OpenAI}, in the summer of 2024 ChatGPT reached 200 millions weekly active users.} with their impressive \textbf{versatility and engaging interfaces that allow verbalizing requests, also for automatic translation} \citep{ouyang2023shifted}. As the MT field explores
such  LLM-based solutions \citep[\textit{inter alia}]{zhu2023multilingual, lyu2024paradigm, alves2024toweropenmultilinguallarge} and integrates MT 
into more complex systems, these solutions have the potential to reshape cross-lingual services 
and user engagement. 

While this transition unfolds, overdue research on the experiences of lay users in cross-lingual and MT settings is gaining urgency. 
%
%
We map this gap and call for first steps to fill it.



%

\section{Three Factors for MT Lay Users} \label{sec:factors}
%

\paragraph{\faHandPointer \space Usability}
The usability of MT systems---how effectively, efficiently, and satisfactorily users can achieve their goals in a given context \citep{ISO9241-11}---is informed and guided by how these systems are evaluated. 
%
The field, however, tends towards performance-driven 
leaderboards 
\citep{rogers2019leaderboards}, which have been criticized for pursuing abstract notions of \emph{accuracy} and \emph{quality} above the practical \emph{utility} of a model or other relevant values \citep{ethayarajh-jurafsky-2020-utility}.
%
These values are often contextual:
\citet{parthasarathi-etal-2021-sometimes-want} discuss how 
\textit{robustness} to misspellings might be detrimental if using MT for learning. 
Also, \textit{faithfulness} is normally 
key to ``MT quality'', but in creative contexts 
like
subtitling, \textit{enjoyability} may take precedence over 
fidelity 
\citep{guerberof2024or}. 

Standard MT metrics 
offer coarse
scores of generic performance
to rank and compare \textit{models}, but are opaque and only 
assume to inform how useful the model is when embedded within the system the user interacts with
\citep{liebling-etal-2022-opportunities}. 
And yet,  
lay people are only involved as 
evaluators to provide  
model-centric insights, rather than to inform their experiences \citep{saldias-fuentes-etal-2022-toward, savoldi-etal-2024-harm}.\footnote{This trend might be exacerbated by AI surrogates,
which have been suggested as a ``replacement'' for human participants \citep{wang2024large, agnew2024illusion}.} Furthermore, 
general-purpose LLMs now confront us with an ``evaluation crisis'' \citep{liao2023rethinking}, where existing methods and predefined benchmarks for modular tasks may be obsolete, failing to capture real-world downstream contexts. This raises the risk of widening the socio-technical gap, where evaluation practices lack validity and might diverge from human requirements in realistic settings.  
%

\paragraph{\faCheck \space Trust} 
To prevent over-reliance on automatic translations, lay users must calibrate an appropriate level of (dis)\textit{trust}. 
Indeed, they risk accepting potentially flawed translations at face value, and trust may be misplaced when an output appears believable but is inaccurate—an issue that is especially harmful in high-stakes contexts \citep{mehandru-etal-2023-physician}.
Prior research on MT has shown that \textit{fluency} and \textit{dialogue flow} can falsely signal reliability \citep{martindale-etal-2021-machine, robertson2022understanding}, and LLMs amplify this issue with their overly confident tone, even when incorrect \citep{xiong2024llmsexpressuncertaintyempirical, kim2024m}. As general-purpose models increasingly replace domain-specific applications, providing mechanisms for trust calibration becomes even more urgent \citep{deng2022generalpurposemachinetranslation, litschko-etal-2023-establishing}.
To harness the benefits of MT systems while avoiding over-reliance on flawed translations, lay users often resort to back-translation\footnote{i.e. automatically translating a text to a target language and then back to the source language.} as a strategy to improve confidence~\citep{back-2007, zouhar-etal-2021-backtranslation, mehandru-etal-2023-physician}. However, back-translation is often performed manually due to the lack of dedicated functionalities, and its soundness
remains debated.
Another critical factor in fostering appropriate trust is \textit{transparency}---e.g. communicating uncertainty and providing explanations~\citep{liao2023aitransparencyagellms}. While explainability work is growing \citep{ferrando2024primer}, ensuring that explanations are informative and digestible to lay users rather than just developers is not trivial. Moreover, how to effectively integrate such uncertainty signals into the development of translation systems and their user interfaces is still an open question.

\paragraph{\faBookOpen \space Literacy}
MT-mediation, as a form of human-machine interaction \citep{green2015natural,o2012translation}, should also regard how lay users themselves play a role in improving interactions and 
apply control strategies to overcome MT limitations.
%
This requires critical agency rather than passive consumption.
In this area, prior work \cite{myabe} has shown that
preventing the display of potentially flawed translations 
causes discomfort to users, indicating that 
they
prefer warnings and guidance over outright blocks. But warnings serve as an initial signal; 
then users
should know how to proceed in recovering from MT errors \citep{shin2013enabling}.
To address this, \citet{bowker2019towards} introduce the concept of MT \textit{literacy}, a digital skill to equip users with the knowledge 
to interact more effectively with MT.\footnote{For online materials, see \url{https://sites.google.com/view/machinetranslationliteracy/}.} This includes pre-editing input text to mitigate common failures (e.g. using short sentences). 
While literacy workshops proved beneficial to students
\citep{Bowker01092020},\footnote{For other data literacy initiatives targeting students, see the DataLitMT project \citep{hackenbuchner-kruger-2023-datalitmt}.}
reaching more vulnerable populations and underserved languages remains a challenge \citep{Liebling}.\footnote{e.g. see BabelDr for a case of MT design for healthcare involving migrant populations: \url{https://babeldr.unige.ch/}.}
Focusing on target 
comprehension, \citet{liebling2021three} explore interfaces with dictionary access and assistive bots.\footnote{See the Lara system, integrating the two-box interfaces with a bot: \url{https://laratranslate.com/translate}.} While LLMs encourage participation through chat and interactive queries \citep{qian2024enabling}, their reliability in this role remains uncertain, as LLM-powered systems may impact cognitive attention required for critical engagement  \citep{zhai2024effects, lee2025impact}. Also, MT literacy must evolve to address new opportunities and failures introduced by LLMs, such as 
cascading errors across multiple requests.

\section{Future Directions and Conclusion} \label{sec:recommendations}
To conclude, we examine directions for future research in 
traditional or LLM-based MT
that integrates lay user perspectives. 
We map such directions and corresponding recommendations to the three factors outlined in Section §\ref{sec:factors}. 

\paragraph{Consider Lay People As Users (\faHandPointer, \faCheck, \faBookOpen)} 
%
To gauge how/when users interact with MT as well as current blindspots
we should consider their experiences rather than just involve them as manual evaluators. Inspired by monolingual work \citep{handaeconomic},  analyzing user logs can help us observe real engagement and preferences. Surveys and \textit{in vivo} research offer qualitative insights into users' perceptions \citep{zheng, robertson2022understanding}. To this aim, it is essential to avoid two main pitfalls: \textit{i)} exploiting participants (see \S\ref{sec:ethics}) and \textit{ii)} treating them as a homogeneous group: 
factors like sociodemographics, education, and stress levels can greatly influence their expectations and interactions \citep{rooein2023know, ai-culture}.

\paragraph{Design for Usability and Utility (\faHandPointer)}
Achieving human-like translations should not be blindly viewed as the ultimate goal---automated text is a means to serve a broader purpose, not an end in itself \citep{caselli-etal-2021-guiding}.
Prior work has evaluated systems based on their success in guiding human decision-making
\citep{zhao-etal-2024-successfully} or by assessing gender bias in MT via user-relevant measures, like time, effort, or economic costs \citep{savoldi-etal-2024-harm}. Research could focus on making measurements more actionable \citep{delobelle-etal-2024-metrics}, e.g. to identify usability thresholds below which MT is no longer beneficial.
Therefore, we should aim to correlate automated approaches with human-centered measurements to harness the benefits of both.\footnote{e.g. replicability and ecological validity, respectively.} 
 However, this is challenging due to the variability of utility values among users and usages. 
Multi-metric and multifaceted approaches like HELM \citep{bommasani2023holistic}
show promise in this area, but future work could further align MT evaluation and design with socio-requirements and prototypical use cases \citep{liao2023rethinking}.

\paragraph{Enrich MT Outputs (\faCheck)}
In user-facing systems, it is crucial to not only focus on generated translations but also to develop methods for \textit{estimating} and \textit{conveying} uncertainty, ambiguities, and errors to ensure reliable usage \citep{xu-etal-2023-understanding,zaranis-etal-2024-analyzing}.
For instance, \citet{briakou-etal-2023-explaining} use contrastive explanations to help users understand cross-linguistic differences, but it is unclear how to disentangle when their approach captures critical errors or simple meaning nuances in the wild. 
Quality estimation can also warn users in real time about flawed translations, though numeric indicators are hard to interpret to lay users \citep{myabe}. Indeed, a key area of future research is how to best communicate digestible information to lay users, e.g. via visualizations.\footnote{e.g. by highlighting errors or reliable keywords.} Textual explanations show promise in communicating uncertainty and avoiding over-reliance in LLMs, but the exact language used is relevant \citep{kim2024m}, and we thus advocate for MT work in this area.

\paragraph{Foster Transparency (\faCheck \space \faBookOpen) and Agency (\faBookOpen)}
%
Users should have the \textit{option} to be active participants when interacting with MT. In addition to real-time explanations, they could receive clear information about MT’s strengths and limitations (e.g.~support across languages). The field might adapt transparency tools like \textit{model cards} \citep{mitchell-2019} into simplified, public-facing versions and support literacy efforts around emerging technologies.\footnote{For example, see the Elements of AI program: \url{https://www.elementsofai.com/}.}
To foster \textit{user agency}---the ability to make informed, intentional decisions about MT use---approaches such as gamification \citep{Chen14122023} could help promote literacy and lightweight critical engagement. Yet, since MT often serves immediate, time-sensitive needs, it remains uncertain whether users always want or are able to engage critically \citep{bucina}. MT experts are well placed to advance these efforts through interdisciplinary collaboration.
\bs{For instance, \citet{yimin2025} investigate how to sustain non-native 
speakers in influencing the production of their message in MT-mediated communication.}

\paragraph{Bridge Interdisciplinary Avenues (\faHandPointer, \faCheck, \faBookOpen)}
Incorporating user needs, values, desiderata, and human factors is still in its early stages in NLP. However, disciplines like human-computer interaction (HCI), experimental psychology, and social sciences have established 
practices to draw from \citep{liao2023rethinking}.
These methodologies may take longer to implement, but they yield useful insights, e.g. on people cognition and trust, or to implement user studies.  
Besides, they offer methods that approximate real-world interactions cost-effectively, e.g. Wizard of Oz tests prior to developing a new method \citep{goyal-etal-2023-else}, or simulating user actions based on past user data \citep{zhang}. These approaches can be highly useful, but---circling back this section---the fundamental first step remains engaging with end users to understand their needs and behaviors first.

\section{Limitations}
\label{sec:limitations}

\paragraph{Factors}

\bs{Our analysis centers on three key criteria.
These are not exhaustive of all user-centered concerns, but they serve as a starting point for a research agenda on human engagement with MT. The selection of these criteria was guided, first, by the aim to capture complementary and distinct dimensions of MT usage, namely: \emph{i)} \textit{usability}---how to align technology with users through model/system adaptation; \emph{ii)} \textit{trust}---how to calibrate user-MT interactions by addressing dynamics of reliance and confidence; and \emph{iii)} \textit{literacy}---how to empower users by fostering their ability to engage with and learn about MT. Second, our choice reflects their recurrence in the literature we reviewed and discussed throughout the paper. At the same time, they resonate with broader debates in adjacent fields: in HCI and translation studies, usability is an established quality characteristic \citep{guerberof2019impact}; in AI governance, the EU AI Act explicitly foregrounds trust/trustworthiness as a core principle\footnote{\url{https://digital-strategy.ec.europa.eu/en/library/ethics-guidelines-trustworthy-ai}.} and introduces literacy as a requirement for fostering awareness and competence.\footnote{\url{https://artificialintelligenceact.eu/article/4/}.}}


\paragraph{Text-to-Text MT} We do not unpack the differences between text-to-text MT and other modalities, such as speech translation and multimodal cross-lingual tasks \citep{papi2025mcif}. While we acknowledge the relevance of these distinctions, we chose to focus on the broadest and most established MT technology. Expanding to other modalities is an important avenue for future work, but our scope was limited by space and focus.

\paragraph{ACL Anthology Query} Our trends assessment of prior work on MT, LLMs, and Users---reported in Figure \ref{fig:trend-overview}---is based on papers published in the ACL Anthology (see Appendix~\ref{app:acl-query}). While including other sources could have further enriched our trend overview, the Anthology remains the main historical reference point in NLP. Hence, it represents an optimal litmus test for assessing trajectories in the field. Still, throughout the paper, we engage with literature from diverse communities, primarily from \textit{translation studies} and \textit{human factors in computing}, to provide a broader interdisciplinary perspective.

\paragraph{Slower Science} Our proposed future directions advocate for user-centered analyses and studies that require more time and resources compared to automated evaluations and \textit{in vitro} experiments, potentially slowing down the research cycle. However, we argue that user-driven insights are crucial and can only yield benefits to align MT with real-world needs and users.

\section{Ethics Statement}
\label{sec:ethics}

In this work, we advocate for user-centered MT research by focusing on lay users. First, unlike human-in-the-loop methods \citep{wang-etal-2021-putting}---which rely on human contributions to enhance model functionality---we prioritize approaches and directions that are intended to serve and benefit users. 

Second, we do not conduct experiments with participants in this paper. Hence, we do not discuss ethical best practices for research in this area, though we deem them as indispensable, e.g. obtaining proper ethical approval, securing informed consent, and ensuring non-intrusive engagement when working with human participants.

Finally, while we broadly discuss lay users, we do recognize that they actually encompass diverse groups and communities.  Many remain underserved by language technologies, particularly speakers of ``low-resource'' languages, and might face well-known biases in NLP tools related to gender \citep{savoldi2025decade}, dialect \citep{blodgett-etal-2020-language}, or social class \citep{curry-etal-2024-classist}. Especially when engaging with more vulnerable communities and user groups, it is important to respect their lived experiences, avoid exploitative research practices, and ensure they are not treated as mere data sources but as valued participants and users---see e.g. \citet{bird-yibarbuk-2024-centering, ramponi-2024-language} and \citet{birhane2022power}.

\section{Acknowledgments}
Beatrice Savoldi is supported by the PNRR project FAIR -  Future AI Research (PE00000013),  under the NRRP MUR program funded by the NextGenerationEU. 
The work presented in this paper is also funded by the Horizon Europe research and innovation programme,  under grant agreement No 101135798, project Meetween (My Personal AI Mediator for Virtual MEETtings BetWEEN People), and the ERC Consolidator Grant No 101086819.

\bibliography{anthology,custom}
\bibliographystyle{acl_natbib}

\newpage

\appendix

\section*{Appendix}

\section{ACL Anthology Query}
\label{app:acl-query}

\noindent To identify research trends in the ACL community (Figure~\ref{fig:trend-overview}), we searched for specific keywords in either the title or abstract of research articles published from \texttt{2015-01-01} to \texttt{2024-12-31} and hosted in the ACL anthology repository.\footnote{\url{https://aclanthology.org} (accessed:~\texttt{2025-02-01}).} Specifically, we use the following keywords in a case-insensitive fashion and including all grammatical numbers by means of regular expressions:
\begin{itemize}
    \item \textbf{machine translation} (\textsc{mt}): \emph{translation}, \emph{machine translation}, \emph{nmt}, and \emph{mt};
    \item \textbf{language models} (\textsc{lm}): \emph{llm}, \emph{language model}, \emph{large language model}, and \emph{foundation model};
    \item \textbf{users} (\textsc{u}): \emph{user}.
\end{itemize}

To reduce noise, we exclude editorials (i.e. those with a \texttt{proceedings} bibtex type) and rare instances of articles without any author from the matching documents. We obtain a total of 62,032 articles, of which 8,072 match \textsc{mt} keywords, 13,977 match \textsc{lm} keywords, and 5,084 match \textsc{u} keywords.

\ar{In Figure~\ref{fig:trend-overview-2}, we present the trends of interest over the last ten years in terms of absolute counts.}

\begin{figure}[!h]
\begin{subfigure}{.24\textwidth}
    \resizebox{\linewidth}{!}{%

    \begin{minipage}[t]{.99\linewidth}
    \centering
    \strut\vspace*{-9mm}\newline
    
        \begin{tikzpicture}
            \begin{axis}[
                /pgf/number format/1000 sep={},
                ylabel={Count},
                ylabel style={yshift=-1mm},
                xmin=2015, xmax=2024,
                ymin=0, ymax=6400,
                xtick={2015,2016,2017,2018,2019,2020,2021,2022,2023,2024,2025},
                ytick={6000,5000,4000,3000,2000,1000,0},
                height=3.6cm,
                width=4.4cm,
                legend pos=north west,
                ymajorgrids=true,
                grid style=dashed,
                title style={yshift=-3mm},
                xlabel style = {font=\footnotesize},
                ylabel style = {font=\footnotesize},
                xticklabel style = {font=\scriptsize, rotate=90},
                yticklabel style = {font=\scriptsize},
                legend style={font=\scriptsize},
            ]

            \addplot[
                color=Salmon!75!white,
                mark=*,
                draw=black!75!white,
                ]
                coordinates {
                (2015,343)(2016,523)(2017,446)(2018,659)(2019,734)(2020,1057)(2021,982)(2022,1089)(2023,1067)(2024,1172)
                };

            \addplot[
                color=teal!75!white,
                mark=*,
                draw=black!75!white,
                ]
                coordinates {
                (2015,34)(2016,97)(2017,96)(2018,179)(2019,355)(2020,804)(2021,1117)(2022,1881)(2023,3045)(2024,6369)
                };

            \addplot[
                color=Salmon!75!white,
                lm-bg halfcircle=90,
                draw=black!75!white,
                ]
                coordinates {
                (2015,5)(2016,20)(2017,11)(2018,32)(2019,48)(2020,92)(2021,98)(2022,149)(2023,228)(2024,493)
                };
            \end{axis}
        \end{tikzpicture}

        \end{minipage}
}%

\end{subfigure}%
\hfill
\begin{subfigure}{.24\textwidth}

    \resizebox{\linewidth}{!}{%

    \begin{minipage}[t]{.99\linewidth}
    \centering
    \strut\vspace*{-9mm}\newline
    
        \begin{tikzpicture}
            \begin{axis}[
                /pgf/number format/1000 sep={},
                ylabel={\textcolor{white}{\%}},
                ylabel style={yshift=-3mm},
                xmin=2015, xmax=2024,
                ymin=0, ymax=1120,
                xtick={2015,2016,2017,2018,2019,2020,2021,2022,2023,2024,2025},
                ytick={1000,800,600,400,200,0},
                height=3.6cm,
                width=4.4cm,
                legend pos=north west,
                ymajorgrids=true,
                grid style=dashed,
                title style={yshift=-3mm},
                xlabel style = {font=\footnotesize},
                ylabel style = {font=\footnotesize},
                xticklabel style = {font=\scriptsize, rotate=90},
                yticklabel style = {font=\scriptsize},
                legend style={font=\scriptsize},
            ]

            \addplot[
                color=Goldenrod!75!white,
                mark=*,
                draw=black!75!white,
                ]
                coordinates {
                (2015,40)(2016,230)(2017,239)(2018,328)(2019,399)(2020,640)(2021,566)(2022,742)(2023,782)(2024,1118)
                };

            \addplot[
                color=Goldenrod!75!white,
                mt-bg halfcircle=270,
                draw=black!75!white,
                ]
                coordinates {
                (2015,5)(2016,31)(2017,21)(2018,29)(2019,30)(2020,52)(2021,63)(2022,63)(2023,78)(2024,73)
                };

            \addplot[
                color=Goldenrod!75!white,
                lm-bg halfcircle=270,
                draw=black!75!white,
                ]
                coordinates {
                (2015,0)(2016,5)(2017,2)(2018,7)(2019,20)(2020,48)(2021,66)(2022,109)(2023,257)(2024,686)
                };
            \end{axis}
        \end{tikzpicture}

        \end{minipage}
}%
\end{subfigure}%

\caption{Trend of interest (absolute counts) in \emph{machine translation} \colorbox{Salmon!75!white}{\textsc{\textbf{mt}}}, \emph{language models} \colorbox{teal!75!white}{\textsc{\textbf{\textcolor{white}{lm}}}}, \emph{users} \colorbox{Goldenrod!75!white}{\textsc{\textbf{u}}}, and combinations thereof in the ACL community over the last 10 years. Besides illustrating the rapid growth of LLM studies, \bs{the left panel highlights the increase in MT research incorporating LLMs (\colorbox{Salmon!75!white}{\textsc{\textbf{mt}}}+\colorbox{teal!75!white}{\textsc{\textbf{\textcolor{white}{lm}}}}), while the right panel shows rising attention to users, particularly in LLM-related work (\colorbox{teal!75!white}{\textsc{\textbf{\textcolor{white}{lm}}}}+\colorbox{Goldenrod!75!white}{\textsc{\textbf{u}}}).}}

\label{fig:trend-overview-2}

\end{figure}

\begin{table*}[t]
    \centering
    \footnotesize
    \begin{tabular} {p{4cm}p{5cm}p{5cm}}
        \toprule
        \textbf{Aspect} & \textbf{Professional Users} & \textbf{Lay Users} \\
        \midrule
        Training & Specialized  & None or limited training \\
        \midrule
        Usage Context & Professional tasks  & Personal or immediate needs \\
        \midrule
        Terminology & Familiar with domain-specific terminology & Can be unfamiliar with specialized terminology translated \\
        \midrule
        Awareness of Limitations & Familiarity with MT capabilities and limitations & Limited/None \\
        \midrule
        Language Proficiency & Proficient in both source and target languages & Limited or no proficiency in at least one of the languages \\
        \midrule
        Error Evaluation & Can effectively judge translation quality and identify errors & May struggle to spot errors \\
        \bottomrule
    \end{tabular}
    \caption{Overview of Professional vs.~Lay Users of Machine Translation.}
    \label{tab:user_comparison}
\end{table*}

\section{Professional and Lay Users of MT}
\label{app:user}

Table~\ref{tab:user_comparison} offers a preliminary outline of some key differences between professional (i.e. translators or MT post-editors) and lay users of machine translation. While we acknowledge that these characteristics often exist along a continuum rather than as clear-cut categories, here we draw on prototypical positions inspired by \citet{user-profile} to highlight contrasting tendencies. This distinction is useful for framing how different user profiles interact with MT systems, particularly in terms of expectations, urgency, error tolerance, and the ability to critically assess output. 

\bs{A more fine-grained subclassification of user types (e.g. according to varying degrees of language proficiency or MT literacy) is currently hindered by the limited literature on these aspects. However, concurrent work by \citet{bassignana-etal-2025-ai} highlights digital literacy as a key potential divide in the use of language technologies. Along these lines, post-editors can develop higher digital literacy and skills through sustained engagement with language technologies as part of their professional growth \citep{secaralt}, in contrast to lay users, whose levels of digital literacy or access can vary greatly.}

\end{document}